\documentclass{article}

\usepackage{microtype}
\usepackage{graphicx}
\usepackage{subfigure}
\usepackage{booktabs}
\usepackage{amsmath}
\usepackage{xcolor}
\usepackage{array}
\usepackage{url}

\usepackage{hyperref}

\usepackage[accepted]{mlsys2025}
\makeatletter\renewcommand{\mlsys@appearing}{Preprint.}\makeatother
\makeatletter\renewcommand{\printAffiliationsAndNotice}[1]{{\let\thefootnote\relax\footnotetext{\hspace*{-\footnotesep}\Notice@String}}}\makeatother
\hypersetup{pdfsubject={}}

\usepackage{enumitem}
\setlist[itemize]{topsep=2pt, partopsep=0pt, itemsep=1pt, parsep=0pt}
\renewcommand{\em}{\it}

\newcommand{\ignore}[1]{}

\newcommand{\boldparagraph}[1]{\vspace*{1ex}\noindent\textit{#1}\hspace{1em}}

\def\cfigure[#1,#2,#3]{
\begin{figure}
\vspace*{0mm}
\begin{center}

\includegraphics[width=3in]{#1} 
 
\vspace*{-3mm}\caption[]{#2
} \label{#3}
 
\vspace*{-5mm}
\end{center}
\end{figure}}

\def\cfigurefour[#1,#2,#3]{
\begin{figure}
\vspace*{0mm}
\begin{center}

\includegraphics[width=4in]{#1} 
 
\vspace*{-3mm}\caption[]{#2
} \label{#3}
 
\vspace*{-5mm}
\end{center}
\end{figure}}

\def\cfiguretemp[#1,#2,#3]{
\begin{figure}
\vspace*{0mm}
\begin{center}

\includegraphics[width=3.5in]{#1} 
 
\vspace*{-3mm}\caption[]{#2
} \label{#3}
 
\vspace*{-5mm}
\end{center}
\vspace*{-2mm}
\end{figure}}

\def\wfigure[#1,#2,#3]{
\begin{figure*}
\vspace*{0mm}
\begin{center}
 \includegraphics[width=\textwidth]{#1} 
 \vspace*{-3mm}\caption[]{#2
} \label{#3}
 
\end{center}
\end{figure*}}

\def\threefigure[#1,#2,#3,#4,#5]{
\begin{figure*}
\vspace*{0mm}
\begin{center}

\begin{tabular}{ccc}
\includegraphics[width=2in]{#1} & \includegraphics[width=2in]{#2} &  \includegraphics[width=2in]{#3} \\
(a) & (b) & (c) \\
\end{tabular}

\vspace*{-3mm}\caption[]{#4
} \label{#5}

\vspace*{-5mm}
\end{center}
\vspace*{-2mm}
\end{figure*}}

\def\dcfigure[#1,#2,#3,#4,#5,#6]{
{
\begin{figure*}
\begin{center}
\begin{minipage}[c]{\columnwidth}{
\includegraphics[width=\columnwidth]{#1} 
\vspace*{0mm}\caption[]{#2} \label{#3} \
}\end{minipage}\hspace*{\columnsep}\
\begin{minipage}[c]{\columnwidth}{
\includegraphics[width=\columnwidth]{#4} 
\vspace*{0mm}\caption[]{#5}\label{#6} \
}\end{minipage}
\end{center}
\end{figure*}
}
}

\def\tableByTable[#1,#2,#3,#4,#5,#6]{
{
\begin{table*}
\begin{center}
\begin{minipage}[c]{3in}{
\centering
{#1}
\vspace*{0mm}\tabcaption[]{#2}\label{#3} \
}\end{minipage}\hspace*{\columnsep}\
\begin{minipage}[c]{3in}{
\centering
{#4}
\vspace*{0mm}\tabcaption[]{#5}\label{#6} \
}\end{minipage}
\end{center}
\end{table*}
}
}

\def\figureByTable[#1,#2,#3,#4,#5,#6]{
{
\begin{figure*}
\begin{center}
\begin{minipage}[c]{3in}{
\centering
\includegraphics[width=\textwidth]{#1}
\vspace*{0mm}\figcaption[]{#2} \label{#3} \
}\end{minipage}\hspace*{\columnsep}\
\begin{minipage}[c]{3.3in}{
\centering
{#4}
\vspace*{0mm}\tabcaption[]{#5}\label{#6} \
}\end{minipage}
\end{center}
\end{figure*}
}
}

\def\tableByFigure[#1,#2,#3,#4,#5,#6]{
{
\begin{figure*}
\begin{center}
\begin{minipage}[c]{4.3in}{
\centering
{#1}
\vspace*{0mm}\tabcaption[]{#2} \label{#3} \
}\end{minipage}\hspace*{\columnsep}\
\begin{minipage}[c]{2.2in}{
\centering
\includegraphics[width=\textwidth]{#4}
\vspace*{-0.35in}\caption[]{#5}\label{#6} \
}\end{minipage}
\end{center}
\end{figure*}
}
}

\def\doublecfigure[#1,#2,#3,#4]{
{
\begin{figure}
\begin{center}
\begin{minipage}[c]{1.5in}{
\begin{center}
\includegraphics[width=1.5in]{#1}
\end{center}
}\end{minipage}\hspace*{1em}\
\begin{minipage}[c]{1.5in}{
\begin{center}
\includegraphics[width=1.5in]{#2}
\end{center}
}\end{minipage}
\vspace*{0mm}\caption[]{#3} \label{#4} \
\end{center}
\end{figure}
}
}

\def\qcfigure[#1,#2,#3,#4,#5,#6]{
{
\begin{figure*}
\vspace*{0.2in}\
\begin{center}
\begin{minipage}[c]{3in}{
\includegraphics[width=3in]{#1} 
\vspace*{-3mm}
}
\end{minipage}\hspace*{0.5in}\
\begin{minipage}[c]{3in}{
\includegraphics[width=3in]{#2} 
\vspace*{-3mm}
}\end{minipage}

\begin{minipage}[c]{3in}{
\includegraphics[width=3in]{#3} 
\vspace*{-3mm}
}
\end{minipage}\hspace*{0.5in}\
\begin{minipage}[c]{3in}{
\includegraphics[width=3in]{#4} 
\vspace*{-3mm}
}\end{minipage}
\end{center}
\caption[]{#5}\label{#6}
\end{figure*}
}
}

\def\twfigure[#1,#2,#3,#4,#5]{
{
\begin{figure*}
\vspace*{0.2in}\
\begin{center}
\begin{minipage}[c]{6.5in}{
\includegraphics[width=6.5in]{#1} 
\vspace*{-3mm}
}
\end{minipage}

\begin{minipage}[c]{6.5in}{
\includegraphics[width=6.5in]{#2} 
\vspace*{-3mm}
}\end{minipage}

\begin{minipage}[c]{6.5in}{
\includegraphics[width=6.5in]{#3} 
\vspace*{-3mm}
}
\end{minipage}
\end{center}
\caption[]{#4}\label{#5}
\end{figure*}
}
}

\def\dwfigure[#1,#2,#3,#4]{
{
\begin{figure*}
\vspace*{0.2in}\
\begin{center}
\begin{minipage}[c]{6.5in}{
\includegraphics[width=6.5in]{#1} 
\vspace*{-3mm}
}
\end{minipage}

\begin{minipage}[c]{6.5in}{
\includegraphics[width=6.5in]{#2} 
\vspace*{-3mm}
}\end{minipage}

\end{center}
\caption[]{#3}\label{#4}
\end{figure*}
}
}

\def\dssfigure[#1,#2,#3,#4,#5,#6]{
{
\begin{figure*}
\vspace*{0.2in}\
\begin{center}
\begin{minipage}[c]{4in}{
\includegraphics[width=4in]{#1}
\vspace*{-3mm}\caption[]{#2} \label{#3} \
}\end{minipage}\hspace*{0.5in}\
\begin{minipage}[c]{2in}{
\includegraphics[width=2in]{#4}
\vspace*{-3mm}\caption[]{#5}\label{#6} \
}\end{minipage}
\end{center}
\vspace*{-0.4in}\
\end{figure*}
}
}

\def\dsfigure[#1,#2,#3,#4,#5,#6]{
{
\begin{figure*}
\vspace*{0.2in}\
\begin{center}
\begin{minipage}[c]{3in}{
\includegraphics[width=3in]{#1}
\vspace*{-3mm}\caption[]{#2} \label{#3} \
}\end{minipage}\hspace*{0.5in}\
\begin{minipage}[c]{3in}{
\hspace*{0.5in}\
\includegraphics[height=3in]{#4}
\vspace*{-3mm}\caption[]{#5}\label{#6} \
}\end{minipage}
\end{center}
\vspace*{-0.4in}\
\end{figure*}
}
}

\def\dsyfigure[#1,#2,#3,#4,#5,#6]{
{
\begin{figure*}
\vspace*{0.2in}\
\begin{center}
\begin{minipage}[c]{2.5in}{
\includegraphics[height=2.5in]{#1}
\vspace*{-3mm}\caption[]{#2} \label{#3} \
}\end{minipage}\hspace*{0.5in}\
\begin{minipage}[c]{2.5in}{
\includegraphics[height=2.5in]{#4}
\vspace*{-3mm}\caption[]{#5}\label{#6} \
}\end{minipage}
\end{center}
\vspace*{-0.4in}\
\end{figure*}
}
}

\def\dyfigure[#1,#2,#3,#4,#5,#6]{
{
\begin{figure*}
\vspace*{0.2in}\
\begin{center}
\begin{minipage}[c]{3in}{
\includegraphics[height=3in]{#1} 
\vspace*{-3mm}\caption[]{#2} \label{#3} \
}\end{minipage}\hspace*{0.5in}\
\begin{minipage}[c]{3in}{
\includegraphics[height=3in]{#4} 
\vspace*{-3mm}\caption[]{#5}\label{#6} \
}\end{minipage}
\end{center}
\vspace*{-0.4in}\
\end{figure*}
}
}

\def\dyoldfigure[#1,#2,#3,#4,#5,#6]{
{
\begin{figure*}
\vspace*{0.2in}\
\begin{center}
\begin{minipage}[c]{3in}{
\epsfysize=2.0in\
\hspace{0.5in}\
\epsfbox{#1}
\vspace*{-3mm}\caption[]{#2} \label{#3} \
}\end{minipage}\hspace*{0.25in}\
\begin{minipage}[c]{3in}{
\epsfysize=2.0in\
\hspace{0.5in}\
\epsfbox{#4}
\vspace*{-3mm}\caption[]{#5}\label{#6} \
}\end{minipage}
\end{center}
\vspace*{-0.4in}\
\end{figure*}
}
}

\def\cfiguredouble[#1,#2,#3,#4]{
\begin{figure}
\vspace*{0.2in}\
\begin{center}
\begin{minipage}[c]{1.5in}{
\epsfxsize=1.5in\
\epsfbox{#1}
}\end{minipage}\hspace*{0.1in}\
\begin{minipage}[c]{1.5in}{
\epsfxsize=1.5in\
\vspace{0.1in}\epsfbox{#2}
}\end{minipage}\vspace*{-0.10in} \caption[]{#3}\label{#4}
\end{center}
\vspace*{-0.4in}\
\end{figure}
}

\def\wpfigure[#1,#2,#3,#4]{
\begin{figure*}
\vspace*{4mm}
\begin{center}

\includegraphics[width=#4]{#1} 

\vspace*{-3mm}\caption[]{#2
} \label{#3}

\vspace*{-5mm}
\end{center}
\end{figure*}}

\def\wprfigure[#1,#2,#3,#4,#5]{
\begin{figure*}
\vspace*{4mm}
\begin{center}

\includegraphics[width=#4, angle=#5]{#1} 

\vspace*{-3mm}\caption[]{#2
} \label{#3}

\vspace*{-5mm}
\end{center}
\end{figure*}}

\def\DoubleFigureWSlide[#1,#2,#3,#4,#5,#6,#7,#8,#9]{
\begin{figure*}
\vspace*{#9}
\begin{center}
\begin{minipage}{#4}
\includegraphics[width=#4]{#1}
\vspace*{-3mm}\caption{#2
}\label{#3}
\end{minipage}
\hspace{2em}
\begin{minipage}{#8}
\includegraphics[width=#8]{#5}
\vspace*{-3mm}\caption{#6
}\label{#7}
\end{minipage}
\vspace*{-5mm}
\end{center}
\end{figure*}
}

\def\DoubleFigureW[#1,#2,#3,#4,#5,#6,#7,#8]{
\begin{figure*}
\vspace*{0in}
\begin{center}
\begin{minipage}{#4}
\includegraphics[width=#4]{#1}
\vspace*{-3mm}\caption{#2
}\label{#3}
\end{minipage}
\hspace{2em}
\begin{minipage}{#8}
\includegraphics[width=#8]{#5}
\vspace*{-3mm}\caption{#6
}\label{#7}
\end{minipage}
\vspace*{-5mm}
\end{center}
\end{figure*}
}

\def\DoubleFigureWHack[#1,#2,#3,#4,#5,#6,#7,#8]{
\begin{figure*}
\vspace*{0in}
\begin{center}
\begin{minipage}{3in}
\includegraphics[width=#4]{#1}
\vspace*{-3mm}\caption{#2
}\label{#3}
\end{minipage}
\hspace{2em}
\begin{minipage}{3in}
\includegraphics[width=#8]{#5}
\vspace*{-3mm}\caption{#6
}\label{#7}
\end{minipage}
\vspace*{-5mm}
\end{center}
\end{figure*}
}

\def\ddcfigure[#1,#2,#3,#4]{
\begin{figure*}
\vspace*{0.2in}\
\begin{center}
\begin{minipage}[c]{\columnwidth}{
\includegraphics[width=\columnwidth]{#1} 
}\end{minipage}\hspace{0.5in}\
\begin{minipage}[c]{\columnwidth}{
\includegraphics[width=\columnwidth]{#2} 
}\end{minipage} \caption[]{#3}\label{#4}
\end{center}
\end{figure*}
}

\def\ddcfigureSlide[#1,#2,#3,#4,#5]{
\begin{figure*}
\vspace*{#5}\
\begin{center}
\begin{minipage}[c]{3in}{
\includegraphics[height=3in]{#1} 
}\end{minipage}\hspace{0.5in}\
\begin{minipage}[c]{3in}{
\includegraphics[height=3in]{#2} 
}\end{minipage}\vspace*{-0.10in} \caption[]{#3}\label{#4}
\end{center}
\vspace*{-0.4in}\
\end{figure*}
}

\def\cxfigure[#1,#2,#3]{
\begin{figure}
\vspace*{4mm}
\begin{center}
 
\epsfxsize=2.5in\
\epsfbox{#1}\
 
\vspace*{-0.10in}\caption[]{#2
} \label{#3}
 
\vspace*{-5mm}
\end{center}
\vspace*{-2mm}
\end{figure}}

\newcommand{\beforecaption}{\vspace{-.15cm}\begin{spacing}{0.85}}
\newcommand{\aftercaption}{\vspace{-.45cm}\end{spacing}}

\newcommand{\sys}{BunshinRL}
\newcommand{\fork}{BunshinFork}

\usepackage{tikz}

\mlsystitlerunning{\sys{}: Efficient Agentic RL Rollouts}

\begin{document}

\twocolumn[
\mlsystitle{\sys{}: Whole-Workspace Forking for Agentic Reinforcement Learning Rollouts}

\begin{mlsysauthorlist}
\mlsysauthor{Yutong Huang}{ucsd}
\mlsysauthor{Vikranth Srivatsa}{ucsd}
\mlsysauthor{Alex Asch}{ucsd}
\mlsysauthor{Yiying Zhang}{ucsd,gensee}
\par\vspace{3pt}
{\small \textsuperscript{1}University of California, San Diego, La Jolla, CA, USA \quad \textsuperscript{2}GenseeAI}
\end{mlsysauthorlist}

\mlsysaffiliation{ucsd}{University of California, San Diego, La Jolla, CA, USA}
\mlsysaffiliation{gensee}{GenseeAI}

\mlsyscorrespondingauthor{Yutong Huang}{yutonghuang@ucsd.edu}

\mlsyskeywords{Agentic Reinforcement Learning, Sandboxes, Copy-on-Write, Memory Management}

\vskip 0.3in

\begin{abstract}
Agentic reinforcement learning (RL) rollouts interleave model reasoning with tool calls in persistent workspaces. Scaling these rollouts incurs substantial memory costs for concurrent environments, while slow tool calls delay entire rollout batches. We present \sys{}, an agentic RL rollout system that exploits shared environment state and recurring tool calls through three techniques: memory sharing across environments, tool-prefix caching, and speculative tool execution. These techniques build on \fork{}, a novel live whole-workspace forking mechanism for containers. \fork{} reconstructs process trees, shares memory copy-on-write, and preserves file, communication, and GUI state, enabling cached-state reuse and rollback after incorrect speculation. We implement \sys{} and \fork{} through Linux kernel module and CRIU modifications. \sys{} reduces peak memory by up to 43\%, total memory consumption by up to 51\%, and end-to-end rollout latency by 23\%.
\end{abstract}

]

\printAffiliationsAndNotice{}

\section{Introduction}
\label{sec:intro}

AI agents are increasingly used to perform computer-operating tasks, with browser agents navigating interactive websites and computer-use agents (CUAs) operating desktop applications through graphical interfaces~\citep{osworld,webarena}. Reinforcement learning (RL) improves their ability to complete multi-step tasks by learning from interaction outcomes~\citep{agenticrlsurvey}. Unlike text-only RL, where rollouts generate responses without interacting with an external environment, agentic RL rollouts interleave model reasoning with tool execution. Each rollout therefore requires a persistent workspace that maintains application and execution state across tool calls and supplies observations for subsequent decisions.

Agentic RL presents two key new challenges.
First, each sample in the training dataset needs an isolated sandbox so that its code execution, file edits, and application interactions cannot corrupt another rollout's state.
A sandbox for a CUA may contain a browser, a display server, desktop applications, and supporting services, all of which consume CPU memory throughout the episode.
With 4\,GB per sandbox, a batch of 256 rollouts requires 1\,TB of CPU memory.
Group-based algorithms such as GRPO further multiply environment instances by sampling several trajectories per task~\citep{deepseekmath}.

Second, tool execution accounts for significant rollout time, as each tool call can take seconds to minutes and there can be a long tool call chain for complex tasks. 
In synchronous RL, a few slow trajectories can delay the completion of an entire rollout batch, not only delaying the RL rollout process but also leaving other resources underutilized.

These costs present two opportunities for reuse. First, environments within a rollout batch can contain substantial identical memory, especially among trajectories sampled from the same task in a GRPO group. Second, the same set of training tasks across iterations can produce tool calls that are exactly the same as seen in previous iterations. Matching tool call prefixes offer an opportunity to skip tool calls in the beginning of trajectories. Recurring calls later in a trajectory provide history from which we can potentially leverage for tool speculative execution.

Based on these insights, we propose \textbf{\sys{}}, an agentic RL rollout system that incorporates three complementary techniques. \emph{Memory sharing} retains one physical copy of common environment state and allocates private pages only as samples diverge in their tool calls. \emph{Tool-prefix caching} restores workspaces that are snapshotted in previous iterations and maintained in memory- and storage-efficient tree structures. \emph{Speculative tool execution} uses a small trained tool selector model to predict what tool calls from previous runs to use for speculative execution. Upon reasoning steps completion, we compare tool call signatures and either keep speculating for matched comparison or roll back to the workspace state before wrong tool speculation. 

Realizing these techniques requires proper support at the systems layer. 
Memory sharing requires runtime copy-on-write (CoW) isolation; prefix reuse must restore the state associated with cached tool results; and incorrect speculation must leave no changes in the original workspace. These requirements motivate a common primitive: \emph{live whole-workspace forking}. A fork shares unchanged parent memory while supporting independent execution. Speculation forks the current workspace before executing a predicted call, whereas caching retains snapshots from which subsequent rollouts can fork.

We realize this primitive with \textbf{\fork{}}, a live forking mechanism for container workspaces. Containers are lightweight and offer the level of isolation that is sufficient for many types of tasks. However, no existing solution supports live container forking. \fork{} forks a container by reconstructing the process tree and execution metadata in isolated namespaces, establishing CoW memory mappings, and providing a private writable file-system view. It also preserves internal communication and GUI state while reconstructing resources that require private copies. 

On top of \fork, we design several optimizations to reduce the latency and memory consumption of \sys. First, we cache only a selected set of workspace snapshots in memory, while leaving the entire snapshots on disk to avoid on-demand disk read and workspace reconstruction latency. Both the in-memory and on-disk snapshots leverage deduplication to save space. Second, we replay some tool calls even when they hit prefix cache, when replaying them is faster than reconstructing snapshots from disk. Third, we only perform speculative tool execution for tail samples, as tail determines rollout latency and speculation incurs some CPU and memory overhead. 

We implement \sys{} from scratch and implement \fork{} by changing the Linux kernel and the CRIU container checkpoint and restore system~\citep{criu}. We evaluate \sys{} by training Qwen3-VL-8B~\citep{qwen3vl} on OSWorld and WebArena, and we
compare it with CRIU and AgentENV~\citep{agentenv}.
On OSWorld, \sys{} reduces peak memory by 30\% compared with CRIU; on WebArena, it reduces peak memory by 43\% compared with AgentENV. It reduces total memory consumption by up to 51\%.
\sys{} improves end-to-end rollout latency by 23\%, while only adding 1\,GB of peak memory and maintaining original reward scores.

Overall, this paper makes the following key contributions:

\begin{itemize}
  \item We identify memory duplication and tool-execution latency as key challenges in agentic RL rollouts.
  \item We propose memory sharing, tool-prefix caching, and speculative tool execution to reduce rollout memory consumption and latency.
  \item We build \sys{}, an integrated agentic RL rollout system that reduces peak memory by up to 43\% and rollout latency by 23\%.
  \item We develop \fork{}, the live whole-workspace forking mechanism underlying \sys{}. It also supports container forking beyond RL, including agent inference.
\end{itemize}

The source code of \sys\ and \fork\ is available at \url{https://github.com/WukLab/ForkRL}.

\section{Motivation}
\label{sec:motivation}

\begin{figure*}[!t]
    \centering
    \input{fig-motivation-shareable-memory}\hfill
    \input{fig-motivation-batch-memory}
\end{figure*}

Efficient agentic RL requires coordinating model generation with many stateful execution environments. We examine the workspace requirements, memory costs, and tool-execution overheads of this workload to motivate the three complementary features of \sys{}: memory sharing, tool-prefix caching, and speculative tool execution (Figure~\ref{fig:use-cases}).

\subsection{Background: Agentic RL}
\label{sec:motivation-background}
Agentic RL trains a policy to complete tasks through interactions with an environment. Training alternates between collecting rollout trajectories and updating the policy using their rewards. Within a trajectory, the policy reasons over the task and previous observations, generates a tool call, and receives an observation from its execution. Group-based algorithms such as GRPO sample multiple trajectories for each task to construct relative reward signals~\citep{deepseekmath}. Unlike text-only RL, collecting agentic RL rollouts requires not only model inference but also environment execution to carry out tool calls and obtain observations.

\subsection{Workspace Environments in Rollout}
\label{sec:motivation-state}
Each rollout needs an isolated workspace whose state persists across tool calls. Systems such as AgentENV~\citep{agentenv} and DSec~\citep{dsec} provide sandbox infrastructure for executing agentic RL rollouts. For a computer-use agent, this workspace may include a browser, desktop applications, a display server, and supporting services. A tool call can modify files, leave processes running, or change application memory and GUI state. 

These properties constrain both reuse and speculation. Reusing a tool result must also recover the workspace state it produced; executing a predicted call must isolate its effects until the prediction is accepted. Copying files alone misses live application state, while restarting and replaying calls adds work and may produce different outcomes because of timing, background activity, or remote responses. Workspace management must therefore account for processes, memory, files, and communication state. External services require separate handling: restoring a local workspace cannot undo effects outside its boundary.

\subsection{Memory Cost in Rollout}
\label{sec:motivation-memory}
A rollout batch requires many concurrent environments, especially when group-based training samples several trajectories for each task. Each environment may hold a large application working set, so maintaining independent copies can quickly exhaust host memory. Yet these environments often contain the same libraries, initialized applications, and task data. Sharing immutable container image layers does not eliminate duplicated process memory accumulated during execution.

Figure~\ref{fig:motivation-shareable-memory} breaks down memory that is identical across environments into file cache, heap, and shared memory. Environments in the same GRPO group exhibit more common state than environments from different tasks, and substantial overlap remains after ten steps. The heap and shared-memory components highlight opportunities beyond sharing immutable files.

Memory sharing can retain one physical copy of common state and allocate private pages only when an environment modifies them. Memory consumption then depends primarily on the shared working set and each trajectory's changes, rather than a full copy for every rollout. Some resources, such as process metadata, still require private storage, and savings decrease as trajectories overwrite shared state. Reducing duplication allows more rollouts to run concurrently under a fixed memory budget and makes room for cached workspace states and speculative executions.

Figure~\ref{fig:motivation-batch-memory} compares peak CPU memory as rollout concurrency increases. The gap between CRIU and \sys{} widens because CRIU maintains private memory copies while \sys{} shares unchanged state. At 256 concurrent rollouts, CRIU uses about 1.8 times as much peak memory, illustrating the capacity benefit of sharing across environments.

\subsection{Tool Calls in Rollout}
\label{sec:motivation-latency}
A rollout spends time both generating model outputs and executing tools. In synchronous training, the batch must wait for its slowest trajectory, so a long browser navigation, test suite, or application operation can delay the next training step even after other rollouts finish. These delays accumulate across tool calls, and faster model inference alone does not remove them.

Repeated training tasks create two opportunities to reduce tool latency. First, trajectories can repeat tool-call prefixes across samples and training iterations. A prefix cache can reuse their observations and resulting workspace states, skipping tool execution while retaining each trajectory's reasoning~\citep{tvcache}. A valid hit must match the originating task, workspace, and tool-call prefix; identical arguments alone are insufficient.

Second, calls that recur after trajectories diverge can inform predictions of upcoming tool invocations. Executing a predicted call alongside policy generation can hide tool latency~\citep{paste}. A matching call can adopt the speculative result and workspace state; a mismatch requires discarding speculative changes and executing the actual call from the prior state. The benefit depends on prediction accuracy, available overlap, and the cost of managing speculative workspaces.

Together, these opportunities motivate \sys{} as an integrated rollout system: memory sharing increases environment concurrency and reduces the cost of retained states, prefix caching avoids repeated tool execution, and speculation overlaps remaining tool work with reasoning. Efficient whole-workspace management enables all three, while reuse must respect the task's execution and sampling semantics.

\section{\sys{}: Agentic Rollout System}
\label{sec:arborrl}

\begin{figure*}[!t]
  \centering
  \includegraphics[width=\textwidth]{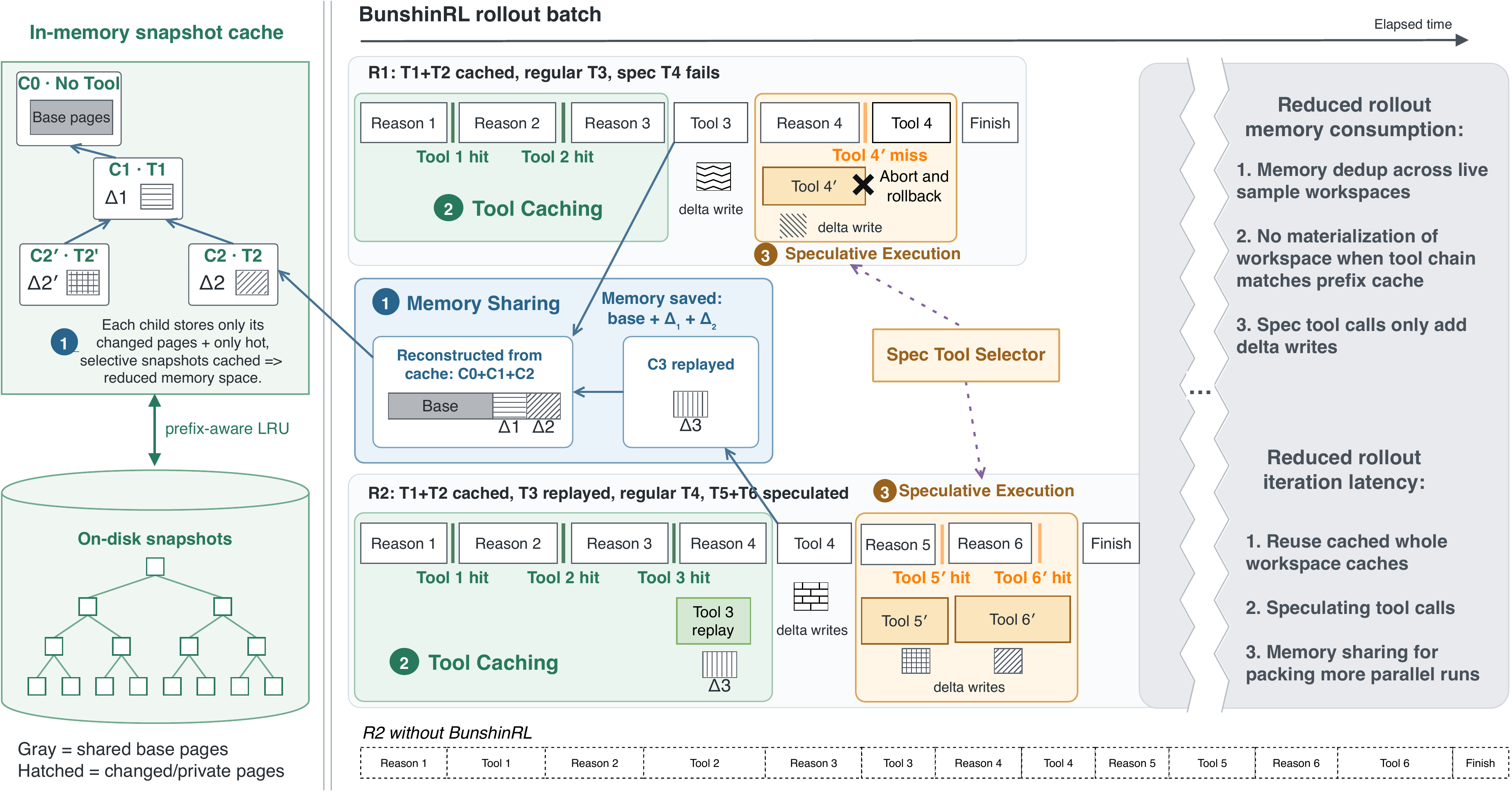}
  \caption{
      \textbf{Illustration of \sys{}.} 
      Right shows two samples in a simplified rollout step; left shows the on-disk and in-memory workspace snapshot stores for tool call prefix. Reasoning steps' generated tool call signatures are matched with the tool prefix tree. All initial matched tool calls are skipped until the first miss in tool signature (Tool 3 for R1 and Tool 4 for R2), at which point the corresponding snapshots are used to reconstruct the workspaces. As C3 is not cached in memory, Tool 3 is replayed for R2 and its delta write is added to the C0+C1+C2 snapshot. The samples execute actual Tool 3 for R1 and actual Tool 4 for R2, creating new delta CoW pages. After these calls finish, speculation starts from the resulting workspaces: Tool 4 for R1 and Tools 5 and 6 for R2. The Spec Tool Selector selects tool calls that are likely to happen from previous tool call history. When a reasoning step finishes, its output tool signature is compared with the speculative tool signature. For failed speculation, the system rolls back to the workspace snapshot before speculation.}
  \label{fig:use-cases}
\end{figure*}

\sys{} manages the execution environments used to collect agentic RL trajectories. It combines three features: memory deduplication across related workspaces, tool-prefix caching that avoids repeated execution, and speculative tool execution that overlaps tool latency with model generation. All three use \fork{}, the whole-workspace forking mechanism to be described in Section~\ref{sec:design}. Figure~\ref{fig:use-cases} illustrates the main components of \sys.


\subsection{Memory Deduplication}
\label{sec:arborrl-memory}

\sys{} reduces the resident memory required by concurrent rollouts by forking workspaces and performing memory copy-on-write (CoW). Rollouts for samples in the training batch begin from one initialized workspace. When they follow a common sequence of tool calls, they reuse cached workspace snapshots (\S\ref{sec:arborrl-cache}) and defer creating a writable branch until their calls diverge. A sample that shares a longer prefix can fork a later snapshot and inherit the state accumulated through that prefix.

\fork{} maps unchanged pages from a source into its descendants with copy-on-write. A branch allocates private pages as it writes, while other descendants continue to reference the shared pages. The memory cost only includes the common memory state, branch-private writes, and reconstruction metadata, rather than a full workspace image for every sample (\S\ref{sec:motivation-memory}).

The same sharing applies inside the prefix snapshot store and to speculative branches. A child snapshot retains changes relative to its parent while referencing inherited pages; a speculative fork adds private state only as it diverges. In Figure~\ref{fig:use-cases}, $C_3$ extends $C_2$ with $\Delta_3$. A sample adopting $C_3$ can share the base, $\Delta_1$, and $\Delta_2$ already mapped by another sample adopting $C_2$. 

\subsection{Tool-Prefix Caching}
\label{sec:arborrl-cache}

\sys{} organizes tool-call prefixes as a tree rooted at the initial workspace state $C_0$. Each edge represents a tool invocation under its parent state. A cache hit requires matching tool call signature prefix, including tool names and arguments at every tool call steps from the initial state. A sample traverses the tree as its policy generates calls. On the first miss (unmatched tool name or argument), \sys{} reconstructs the cached state (up until the last tool hit) and then uses \fork{} to create a writable branch. 

\boldparagraph{Snapshot cache management.}
We record all previous iterations' workspace states after every tool call for all samples. Storing the raw snapshots of all of them will consume huge amounts of storage space. We leverage a similar deduplication idea from \S\ref{sec:arborrl-memory} by storing only the deltas of what a tool call changes at every child node of a prefix tree, as shown in Figure~\ref{fig:use-cases}. The entire prefix tree is then stored on disk. Upon prefix matching, we restore the workspace by reading from the root to the last matched tool call and add deltas of every tree node to the base snapshot. Although this technique makes the snapshot tree small enough to fit on disk, reading tree nodes from disk and reconstructing workspaces can be slow and offset the latency saved by prefix reuse.
\sys{} therefore retains hot workspace snapshots in memory. We design an ancestor-aware LRU policy which maintains an ancestor-closed cache: accessing a snapshot refreshes its recency and retains all its snapshot ancestors in memory. Eviction preserves this invariant, so an ancestor remains resident while any cached descendant depends on it. 

\boldparagraph{Sparse snapshots and replay.}
To further reduce memory consumption caused by snapshot caches, \sys{} materializes snapshots at selected tool-call boundaries rather than every step, while retaining tool-call metadata. A prefix without its own snapshot is reconstructed by forking the nearest retained ancestor and replaying the intervening calls. The replay overlaps with model reasoning and thus keeps reconstruction off the critical path, e.g. Tool 3 replay with Reason 4 for R2 in Figure~\ref{fig:use-cases}. Our sparse snapshot mechanism trades additional tool execution for a smaller memory footprint without adding latency.

Overall, the memory footprint introduced by snapshots is much smaller than the memory saving we get from deduplication (\S\ref{sec:arborrl-memory}) but largely speeds up the rollout phase.


\subsection{Speculative Tool Execution}
\label{sec:arborrl-spec}

We propose speculative tool execution in \sys{} to hide tool latency behind model reasoning. Our key insight is that recurring tool calls performed in earlier training iterations provide a basis for predicting future actions in later iterations. 

\boldparagraph{Speculation predictor and mechanism.}
We train a small classifier model, \textit{Spec Tool Selector}, which predicts the tool call from all historically recorded tool calls based on the sample's trajectory so far. \sys{} forks the current workspace and executes predicted calls in parallel with reasoning.
As soon as the policy produces the actual signature, \sys{} compares it with the prediction. A match adopts the speculative observation and state once execution completes and its dependencies are validated; an unfinished call incurs only its remaining execution time. A mismatch immediately aborts the speculative call, discards its private writes, and starts the actual call from the preceding valid snapshot. Figure~\ref{fig:use-cases} shows failed speculation for $R_1$ and successful speculation for $R_2$.

To further overlap tool-call speculation with model reasoning, we predict $K$ number of tool calls at a time and execute them in the speculation workspace. Each speculated tool call is compared with actual tool signatures as soon as the corresponding reasoning step finishes. The first mismatch aborts the whole speculation chain.

\boldparagraph{Tail-aware selection.}
Even when we perform memory-deduplicated fork for speculation execution, it unavoidably still consumes some additional CPU and memory. To reduce this overhead, we perform speculation only to samples causing the long tail in a rollout batch, as that is what determines the rollout iteration's total latency. We use recorded timings from previous training iterations to identify historically slow samples and long tool calls, and we only perform speculation for those samples. 
This selection concentrates speculative work on the most valuable opportunities; other samples use ordinary tool execution and prefix caching.

Overall, tool call speculation reduces overall rollout latency while adding only tiny memory overhead thanks to our whole-workspace forking and memory deduplication mechanisms.
\section{\fork{}: Workspace Fork}
\label{sec:design}

This section describes \fork{}, the whole-workspace forking mechanism underlying \sys{} (\S\ref{sec:arborrl}), and how it preserves isolated execution while sharing unchanged state.

\subsection{Why Fork Containers}
\label{sec:design-why}

Agentic RL maintains rollout environments, cached prefixes, and speculative branches whose memory footprint limits concurrency. \fork{} uses containers so the host kernel can directly share anonymous memory copy-on-write and unchanged file-backed pages through its page cache, without duplicating a guest kernel for each environment. This reduces the memory cost of the three optimizations in \sys{}. Fork creation must also be fast: its overhead can otherwise offset the latency saved by prefix reuse and speculative execution.

\begin{figure*}[t]
  \centering
  \includegraphics[width=0.8\textwidth]{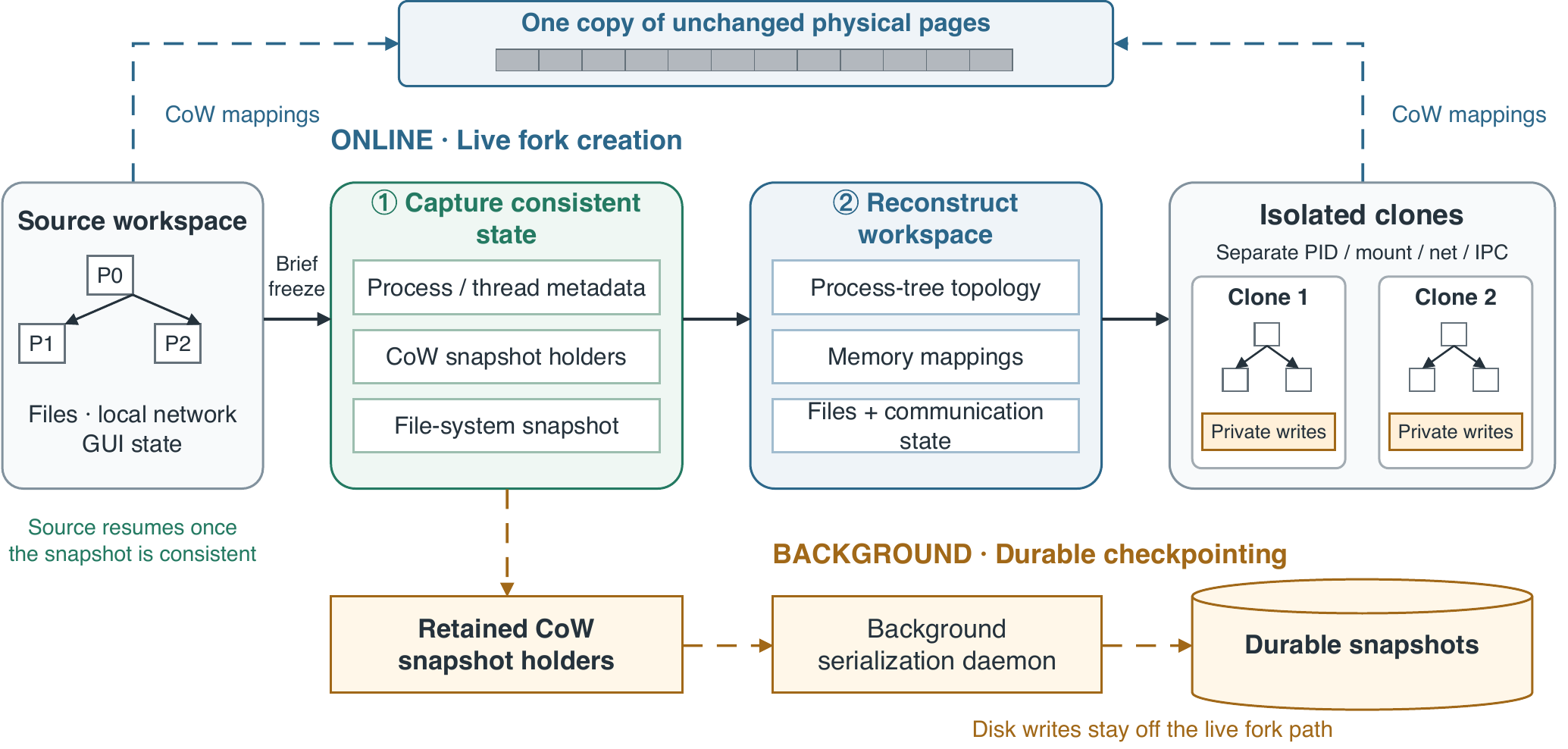}
  \vspace{-6pt}
  \caption{\textbf{\fork{} architecture.} A brief freeze captures a consistent workspace state. The live fork path reconstructs isolated clones that share unchanged pages copy-on-write and allocate private pages on writes. Retained snapshot holders support asynchronous serialization, allowing the source to resume without waiting for disk writes.}
  \label{fig:arborfork-architecture}
\end{figure*}

MicroVMs provide a stronger isolation boundary but introduce guest-kernel and memory-management overheads~\citep{dsec}. Container isolation is sufficient for our web and GUI rollout tasks, which require no guest-level privileges. Workloads requiring privileged operations, such as nested containers, may instead favor VMs~\citep{kimik3,dsec}.

\begin{figure*}[t]
    \centering
    \includegraphics[width=0.8\linewidth]{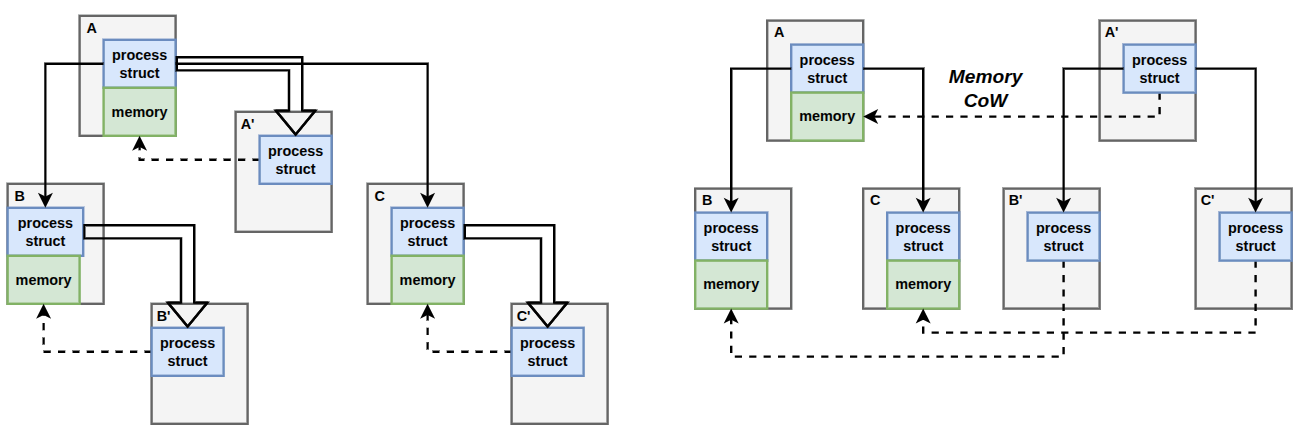}
    \caption{
    \textbf{Linux Process Fork vs. \fork\ Process-Tree Fork}. Left: native \texttt{fork()} duplicates one process as a child of the source process. Right: \fork{} creates a sibling container and reconstructs the whole process tree inside an independent namespace.
    }
    \label{fig:process-fork}
\end{figure*}

\subsection{Fork Overview}
\label{sec:design-overview}

\fork{} captures a consistent workspace boundary and reconstructs its processes, memory mappings, files, and communication resources in an isolated container. Its central principle is to separate \emph{online fork creation from durable checkpointing}: unchanged memory is shared rather than eagerly copied, while serialization to storage proceeds asynchronously. The source resumes once the logical snapshot is consistent, allowing rollouts to continue without waiting for disk writes.

To fork the entire process tree, \fork\ briefly freezes the source and records its process hierarchy, threads, open descriptors, signal and synchronization state, and the register state needed to resume execution. It then creates a destination with independent namespaces and restores the recorded topology, preserving application-visible process identifiers and relationships while assigning distinct host identifiers. Separate PID, mount, network, and IPC namespaces isolate the restored resources from those of the source. Corresponding source and destination processes are linked for memory sharing. A single frozen source can supply multiple clones, which subsequently execute independently.

The freeze covers metadata capture, memory snapshot-holder creation (\S\ref{sec:design-mem}), and a coordinated file-system snapshot. It copies metadata and page tables rather than the application working set. Once these components describe the same point in execution, the source resumes and any durable serialization continues in the background. 
Figure~\ref{fig:arborfork-architecture} illustrates the key mechanisms of \fork{}.

\subsection{Process Subsystem}
\label{sec:design-proc}


Forking each process independently does not reproduce a workspace: Linux \texttt{fork()} makes each new process a child of its source, breaking the original parent--child relationships. As Figure~\ref{fig:process-fork} illustrates, \fork{} instead reconstructs the original process-tree topology inside an isolated container. 
To do so, \fork{} reads the parent--child relationships from the frozen snapshot and recreates them among the processes in the new container.

\subsection{Memory Subsystem}
\label{sec:design-mem}

Rollouts and speculative branches inherit application state and modify it incrementally. Eagerly copying that state would increase both fork latency and memory consumption. \fork{} therefore shares inherited memory copy-on-write (CoW), allocating private pages only as branches diverge.

\fork{} distinguishes anonymous memory from file-backed and shared memory. Anonymous memory has no backing file; file-backed memory includes executable images, libraries, and mapped application data. Shared-memory regions support communication among processes and must preserve those relationships within each restored workspace.

Today's Linux \texttt{fork()} establishes sharing between a parent and its child within a workspace, but \fork{} must share memory between corresponding processes in separate containers. To achieve this, we build a Linux kernel module to install write-protected mappings of the source's anonymous pages in each destination without copying their contents. The kernel's existing CoW fault handling then allocates a private page on the first write by either side, leaving other branches' mappings unchanged. Both the source and its clones can therefore continue writing without changing the state observed by another rollout. Multiple clones can share the same source pages, so their memory cost follows private writes rather than full workspace copies.



\subsection{File-System Subsystem}
\label{sec:design-fs}

\fork{} combines a read-only btrfs snapshot with a private overlayfs layer for each clone. The snapshot is taken while the source is frozen; clones read unchanged files from the shared lower layer and write to their own upper layers. Because siblings read the same underlying files, the host page cache can retain one copy of their unchanged pages.

When a clone is forked again, reflink copies of its upper layer preserve its modifications while sharing existing disk blocks. Overlayfs copies a file into the private upper layer on its first write, so modifying a large file can still incur substantial copy-up overhead.

\subsection{Network Subsystem}
\label{sec:design-net}

\fork{} distinguishes connections within a workspace from connections to external services. Internal connections, such as those between a browser and a local server, are part of the captured state. Their TCP state is restored using kernel TCP-repair mechanisms inside each clone's independent network namespace, avoiding endpoint and port conflicts.

External connections are closed during fork and re-established independently by each clone: a remote server cannot safely treat multiple branches as the same endpoint. Duplicating a live connection would leave ownership and message ordering ambiguous and could replay operations with externally visible side effects. Speculative branches can restrict external networking to prevent uncommitted tool calls from producing remote effects. Restoring a local snapshot cannot undo effects outside the workspace.


\subsection{GUI Subsystem}
\label{sec:design-gui}

A containerized application that relies on a host X11~\cite{X11} or Wayland~\cite{Wayland} server leaves window, input, clipboard, and compositor state outside the workspace. Restoring the application alone can therefore produce an inconsistent GUI.

\sys{} instead runs the GUI stack inside each workspace using Webtop~\cite{webtop} and Selkies~\cite{Selkies}. The Wayland compositor and its clients reside together in the container, with the desktop exported over HTTPS. Keeping this state together ensures that a restored agent observes the desktop associated with its restored applications, rather than a separate host session. \fork{} reconstructs them together, giving each rollout or speculative branch an independent display session, cursor, windows, clipboard, and input stream. 

\section{Evaluation}
\label{sec:evaluation}

\begin{figure*}[t]
    \centering
    \input{fig-e2e-memory-osworld}\hfill
    \input{fig-e2e-memory-webarena}\hfill
    \input{fig-e2e-throughput}
\end{figure*}

We implemented \sys{} from scratch; the rollout algorithm is based on GRPO, and the rollout phase runs on vLLM~\cite{vllm-git-repo}. We implemented \fork{} with the Linux kernel module and extended the CRIU system. 
We run all experiments on two servers: a GPU server with one NVIDIA H100 GPU serves the policy model during rollouts and trains it synchronously; a workspace server hosts all sample workspaces and stores snapshot caches.
The latter has two Intel Xeon Gold 5218 CPUs with 64 hardware threads in total, 256\,GB of DDR4 memory, and a 512\,GB NVMe SSD.

\subsection{Experimental Setups and Baselines}

We evaluated \sys{} using OSWorld~\citep{osworld} and WebArena~\citep{webarena}.
OSWorld has 369 desktop tasks that use applications such as a web browser,
an office suite, a media player, and a combination of common desktop based CUA tasks.
WebArena has web tasks on several self-hosted sites, and we selected its shopping site tasks subset as our training workload.
Each WebArena environment runs the shopping site and its database in the same container as the browser. We train Qwen3-VL-8B~\citep{qwen3vl} with a simple synchronous GRPO loop~\citep{deepseekmath} on a subset of these two workloads.

\boldparagraph{Baselines.}
We compare \sys{} with two systems that fork environments, using the same GRPO loop, policy, and tasks.
\textbf{CRIU}~\citep{criu} forks a container by writing a checkpoint to disk and restoring each fork from it.
Each restored fork holds a private copy of the source's memory.
\textbf{AgentENV}~\citep{agentenv} is a sandbox system designed for the Kimi K3 RL environment. It builds a fork mechanism on top of Firecracker microVM~\citep{firecracker}. 
We give AgentENV the same workspace images as \sys{} and CRIU.

\subsection{Rollout Memory}
\label{sec:rollout-memory}

Figures~\ref{fig:e2e-memory-osworld} and~\ref{fig:e2e-memory-webarena} show the memory of all environments on the
environment server over one training epoch.
Memory rises as rollouts start, peaks while most rollouts run at once, and falls as rollouts finish.
On OSWorld, \sys{} reduces peak memory by 30\% compared with CRIU and by 26\% compared with AgentENV.
On WebArena, \sys{} reduces peak memory by 43\% compared with AgentENV.

\boldparagraph{Peak memory.}
Containers share the host kernel, so the host kernel sees every page of every environment.
\fork{} uses this to share exactly the pages that a fork has not written (\S\ref{sec:design-mem}).
Forks also read unchanged files through one copy in the host page cache (\S\ref{sec:design-fs}).
CRIU restores every fork with a private copy of the source's memory, so each fork adds the memory of a whole
environment.
AgentENV forks share the snapshot's memory copy-on-write, so AgentENV uses less memory than CRIU at the same batch
size.
AgentENV also shares file data through the host page cache and returns free guest pages to the host.
However, each microVM runs its own guest kernel, whose page tables and other metadata stay private to the microVM.
A microVM also needs two levels of page tables, the guest kernel's and the host's for guest memory, while a container
needs only the host's.

\boldparagraph{Epoch length.}
All three systems replay the same recorded trajectories, so differences in epoch length come from the environments.
On OSWorld, CRIU takes 1.3$\times$ longer than \sys{} to finish the epoch.
Each CRIU fork writes a checkpoint to disk and restores a private copy of memory, so forking takes longer.
On WebArena, a full batch of CRIU environments does not fit in the server's 256\,GB of memory and runs out of
memory, so we lower CRIU's batch size.
Smaller batches and slower forks together make CRIU take 2.4$\times$ longer than \sys{} to finish the epoch.
Its peak memory is lower than AgentENV's only because it runs fewer environments at once.
Even so, \sys{} runs the full batch with 36\% less peak memory than CRIU.
This shows that memory also limits rollout throughput, because environments that do not fit force a smaller batch.

AgentENV forks quickly, but its OSWorld epoch still takes 1.2$\times$ longer than \sys{}'s, and its memory stays near
its peak for over half the epoch.
An OSWorld environment with Chrome runs about 40 processes, and a batch runs many environments at once on 64 hardware
threads, so the CPU is heavily oversubscribed.
In a container, the host kernel sees every process and can run it on any free CPU.
In a microVM, the host schedules virtual CPUs and the guest kernel schedules processes on them, so every process is
scheduled twice.
Each switch between virtual CPUs is also a VM context switch, with a VM exit and a VM entry.
A microVM's processes can also run only on its own virtual CPUs, even when other host CPUs are idle.
As a result, tool calls take longer in AgentENV than in \sys{}, and AgentENV's environments hold their memory longer.

\subsection{Tool-Prefix Caching and Speculative Tool Execution}
\label{sec:eval-prefix-spec}

\begin{figure*}[t]
    \centering
    \input{fig-e2e-prefix-spec-memory}\hfill
    \input{fig-ablation-hit-rate}\hfill
    \input{fig-ablation-rewards}
\end{figure*}


All experiments in this subsection run on OSWorld and compare three configurations of \sys{}.
\emph{Base} \sys{} shares memory across forks but uses neither tool-prefix caching nor speculative tool execution.
\emph{Spec}, labeled Spec tool call in the figures, adds speculative tool execution to Base.
\emph{Full} \sys{} adds both tool-prefix caching and speculative tool execution to Base.
We run one epoch with each configuration, and we also train with each configuration for five epochs.

\boldparagraph{Rollout latency.}
In synchronous RL training, a step waits for its slowest rollout.
A few stragglers therefore set the step time and leave the GPU underused while it waits for them.
Rollout latency matters both overall and at the tail, and the two techniques target these two parts.
Tool-prefix caching skips the tool calls that match a cached prefix, so it shortens many rollouts and lowers latency
overall.
Speculative tool execution runs a predicted tool call while the policy is still generating, so it hides the latency of
slow tool calls and helps most for the stragglers.

Figure~\ref{fig:e2e-throughput} shows how many rollouts are still running over the epoch, and the epoch ends when the
last rollout finishes.
Base finishes the epoch in 20.8 minutes.
Spec finishes it in 19.2 minutes, which is 1.1$\times$ faster, and Full finishes it in 17.0 minutes, which is
1.2$\times$ faster.
Spec mainly shortens the tail.
Half of its rollouts finish only 0.5 minutes earlier than in Base, but its last rollout finishes 1.6 minutes earlier.
Full adds prefix caching and finishes rollouts earlier throughout the epoch.
Half of its rollouts finish 1.4 minutes earlier than in Base, and its last rollout finishes 3.8 minutes earlier.
Prefix caching skips matched calls entirely, while speculation hides only the part of a tool call that overlaps with
generation, and a misprediction still runs the actual call afterward.

\boldparagraph{Memory cost.}
Keeping prefix snapshots and speculative forks adds little memory, because each one is a copy-on-write fork.
A cached snapshot starts with no private copy of its anonymous memory and shares all of it with the environment it
came from.
The snapshot is frozen and never writes, so it holds a page of its own only when that environment later writes to the
page.
Its cost therefore grows with how much the environment changes after the snapshot, not with the size of the
environment.
Snapshots along one prefix also share the pages they inherit from each other (\S\ref{sec:arborrl-memory}).
Beyond these pages, each snapshot or fork holds only its own copy of shared-memory regions, such as display buffers,
and its own page tables.
A mispredicted speculative fork is discarded as soon as the policy model issues a different call, so it holds its memory
only briefly.
As Figure~\ref{fig:e2e-prefix-spec-memory} shows, Full and Spec raise peak memory by only about 1\,GiB over Base, from
69\,GiB to 70\,GiB.
Full also lowers memory late in the epoch, because its rollouts finish sooner and release their environments.

\boldparagraph{Hit rate.}
Figure~\ref{fig:ablation-hit-rate} shows the share of tool calls in each training epoch that are served from a cached
snapshot or a speculative fork instead of running on the rollout's critical path.
Full starts at 40\% in the first epoch, where hits come from rollouts of the same GRPO group that share a prefix.
Its hit rate rises to 65\% by the third epoch, because the cache now holds snapshots from earlier epochs.
Spec stays between 41\% and 53\%.
From the second epoch on, Full hits 10 to 13 points more often than Spec, because prefix-cache hits add to
speculation hits.
Both hit rates drop in the fifth epoch as model weights drift more significantly and it starts to issue calls that earlier epochs did
not make.

\boldparagraph{Reward.}
Tool-prefix caching and speculative tool execution do not change the reward.
Figure~\ref{fig:ablation-rewards} shows the average reward in each epoch of the same three training runs.
The three configurations reach the same reward in every epoch.

\subsection{Fork Microbenchmarks}
\label{sec:eval-micro}

\begin{figure}[t]
    \centering
    \includegraphics[width=0.65\linewidth]{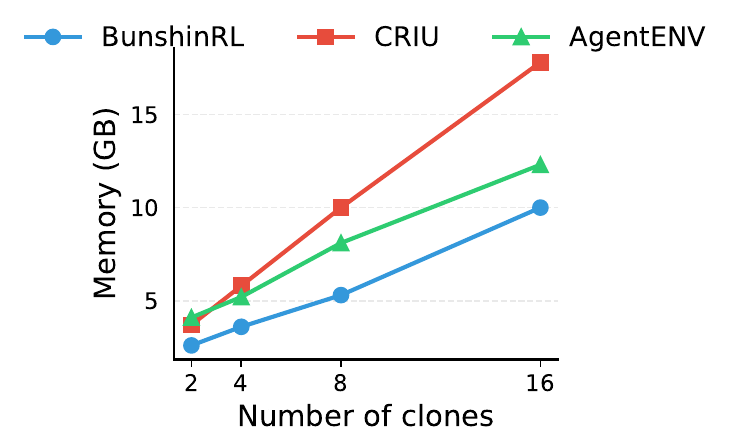}
    \caption{
    \textbf{Memory of Forks.} Total memory of a source environment and its forks as the number of forks grows, for
    \fork{}, CRIU, and AgentENV.
    }
    \vspace{-0.1in}
    \label{fig:ablation-clone-memory}
\end{figure}

\boldparagraph{Fork memory.}
Figure~\ref{fig:ablation-clone-memory} forks one OSWorld environment 2 to 16 times and measures the total memory of the
source and its forks.
At 16 forks, \sys{} uses 10.0\,GB, while CRIU uses 17.8\,GB and AgentENV uses 12.3\,GB.
CRIU adds about 1.0\,GB per fork, because each restored fork holds a private copy of the source's process memory.
\sys{} adds about 0.5\,GB per fork, which is the memory a fork cannot share, such as the display buffers that \fork{}
keeps private (\S\ref{sec:design-mem}) and the kernel metadata of the fork's processes.
AgentENV adds about as much per fork as \sys{}, because its forks also share the snapshot's memory copy-on-write.
However, AgentENV starts higher, and it uses 1.5\,GB to 2.8\,GB more than \sys{} at every fork count.

\section{Related Work}
\label{sec:related}

\boldparagraph{Agentic RL rollout systems.}
Seer, VenusRL, and tree-structured RL optimize scheduling, template sharing, and trajectory branching~\citep{seer,venusrl,treegrpo}. \sys{} instead exploits live workspace state at intermediate rollout steps, combining memory sharing, tool-prefix caching, and tool speculation.

\boldparagraph{Sandbox infrastructure.}
AgentENV supports Firecracker snapshots and live-environment forks~\citep{agentenv,kimik3}; DSec optimizes sandbox provisioning, image sharing, and memory reclamation~\citep{dsec}. \fork{} branches containers through direct CoW sharing of live application memory, without first serializing it. This complements cluster-level provisioning and scheduling.

\boldparagraph{Checkpointing and branchable workspaces.}
CRIU, DeltaBox, Crab, and StateFork preserve execution state through checkpoints and layered storage~\citep{criu,deltabox,crab,statefork}. Our distinction is not state preservation alone: \fork{} directly shares unchanged live memory across isolated process trees, reducing concurrent-branch memory costs while persisting snapshots asynchronously.

\boldparagraph{Tool caching and environment reuse.}
TVCache reuses tool values and associated state; its Docker backend preserves files and selected shell metadata~\citep{tvcache}. ToolCaching manages tool responses, while tree-search agents can reconstruct prefixes by replay~\citep{toolcaching,rmcts}. \sys{} caches resumable workspaces, including running processes and memory, so subsequent actions inherit the corresponding application state. Unlike CUA-Sandbox's runtime/application separation~\citep{cuasandbox}, \fork{} requires no application-specific decomposition. Reuse must still respect sampling semantics: shared stochastic outcomes can alter policy updates~\citep{gupta2026cache}.

\boldparagraph{Speculative tool execution.}
PASTE, draft-model tool prediction, and search speculation overlap predicted calls with generation~\citep{paste,nichols2025speculative,huang2025searchspec}. Speculative Actions also considers sandboxing and rollback~\citep{speculativeactions}. \sys{} contributes a concrete live CoW runtime for mutable workspaces: matching predictions retain both results and application state; mismatches discard isolated branches. External effects still require confinement or deferral.

\boldparagraph{Serverless memory sharing.}
Catalyzer, SEUSS, and Mitosis accelerate sandbox creation through shared execution state~\citep{catalyzer,cadden2020seuss,mitosis}; Medes and KSM deduplicate existing pages~\citep{medes,ksm}. \fork{} targets intermediate states of long-running, multi-process agent workspaces, enabling \sys{} to combine memory sharing with tool reuse and speculation in RL rollout.

\section{Conclusion}
\label{sec:conclusion}

We presented \sys{}, an agentic RL rollout system, and \fork{}, its live whole-workspace forking mechanism. Through memory sharing, tool-prefix caching, and speculative tool execution, \sys{} reduces peak memory by up to 43\% and rollout latency by 23\%. Together, these techniques make stateful agent training more resource-efficient and provide a reusable foundation for branching execution in other agent workloads.

\bibliography{sysml,all-defs,references}

@string{apr = "April" }

@string{jun = "June" }

@string{computer= "IEEE Computer" }

@inproceedings{sheng2025hybridflow,
  title     = {HybridFlow: A Flexible and Efficient RLHF Framework},
  author    = {Sheng, Guangming and Zhang, Chi and Ye, Zilingfeng and Wu, Xibin and Zhang, Wang and Zhang, Ru and Peng, Yanghua and Lin, Haibin and Wu, Chuan},
  booktitle = {Proceedings of the Twentieth European Conference on Computer Systems},
  series    = {EuroSys '25},
  pages     = {1279--1297},
  year      = {2025},
  publisher = {Association for Computing Machinery},
  address   = {New York, NY, USA},
  doi       = {10.1145/3689031.3696075},
  url       = {https://doi.org/10.1145/3689031.3696075}
}

@article{osworld,
  author        = {Tianbao Xie and Danyang Zhang and Jixuan Chen and Xiaochuan Li and Siheng Zhao and Ruisheng Cao and Toh Jing Hua and Zhoujun Cheng and Dongchan Shin and Fangyu Lei and Yitao Liu and Yiheng Xu and Shuyan Zhou and Silvio Savarese and Caiming Xiong and Victor Zhong and Tao Yu},
  title         = {{OSWorld}: Benchmarking Multimodal Agents for Open-Ended Tasks in Real Computer Environments},
  journal       = {arXiv preprint arXiv:2404.07972},
  year          = {2024},
  doi           = {10.48550/arXiv.2404.07972},
  url           = {https://arxiv.org/abs/2404.07972}
}

@article{X11,
  author  = {Robert W. Scheifler and Jim Gettys},
  title   = {The {X} Window System},
  journal = {ACM Transactions on Graphics},
  volume  = {5},
  number  = {2},
  pages   = {79--109},
  month   = apr,
  year    = {1986},
  doi     = {10.1145/22949.24053}
}

@manual{Wayland,
  author       = {Kristian Hoegsberg},
  title        = {The {Wayland} Protocol},
  organization = {Wayland Project},
  year         = {2012},
  url          = {https://wayland.freedesktop.org/docs/html/},
  note         = {Accessed: 2026-05-15}
}

@misc{webtop,
  author       = {{LinuxServer.io}},
  title        = {{Webtop}: Linux in a Web Browser},
  year         = {2026},
  howpublished = {\url{https://github.com/linuxserver/docker-webtop}},
  note         = {Accessed: 2026-05-15}
}

@misc{Selkies,
  author       = {{Selkies Project}},
  title        = {{Selkies}: Linux {WebRTC} {HTML5} Remote Desktop Streaming Platform},
  year         = {2024},
  howpublished = {\url{https://github.com/selkies-project/selkies}},
  note         = {Accessed: 2026-05-15}
}

@misc{criu,
  author       = {{Checkpoint/Restore Project}},
  title        = {{CRIU}: Checkpoint/Restore in Userspace},
  year         = {2025},
  howpublished = {\url{https://criu.org/Main_Page}},
  note         = {Accessed: 2026-05-15}
}

@inproceedings{firecracker,
  author    = {Alexandru Agache and Marc Brooker and Andreea Florescu and Alexandra Iordache and Anthony Liguori and Rolf Neugebauer and Phil Piwonka and Diana-Maria Popa},
  title     = {Firecracker: Lightweight Virtualization for Serverless Applications},
  booktitle = {17th USENIX Symposium on Networked Systems Design and Implementation (NSDI 20)},
  year      = {2020},
  pages     = {419--434},
  address   = {Santa Clara, CA},
  publisher = {USENIX Association},
  isbn      = {978-1-939133-13-7},
  url       = {https://www.usenix.org/conference/nsdi20/presentation/agache}
}

@inproceedings{cadden2020seuss,
  title={{SEUSS}: skip redundant paths to make serverless fast},
  author={Cadden, James and Unger, Thomas and Awad, Yara and Dong, Han and Krieger, Orran and Appavoo, Jonathan},
  booktitle={Proceedings of the Fifteenth European Conference on Computer Systems},
  pages={1--15},
  year={2020}
}

@misc{vllm-git-repo,
  title={vLLM: Easy, Fast, and Cheap LLM Serving with PagedAttention},
  author={{vLLM Team}},
  howpublished={\url{https://github.com/vllm-project/vllm}},
  year={2023},
  note={GitHub repository}
}

@article{kimik3,
  author        = {{Kimi Team}},
  title         = {Kimi {K3}: Open Frontier Intelligence},
  journal       = {arXiv preprint arXiv:2607.24653},
  year          = {2026},
  url           = {https://arxiv.org/abs/2607.24653}
}

@article{dsec,
  author        = {Jialiang Huang and Hongxuan Tang and Jingchang Chen and others},
  title         = {{DeepSeek Elastic Compute (DSec)}: A Sandbox Infrastructure for Effective Agentic Training at Scale},
  journal       = {arXiv preprint arXiv:2609.22978},
  year          = {2026},
  url           = {https://arxiv.org/abs/2609.22978}
}

@inproceedings{webarena,
  author        = {Shuyan Zhou and Frank F. Xu and Hao Zhu and Xuhui Zhou and Robert Lo and Abishek Sridhar and Xianyi Cheng and Tianyue Ou and Yonatan Bisk and Daniel Fried and Uri Alon and Graham Neubig},
  title         = {{WebArena}: A Realistic Web Environment for Building Autonomous Agents},
  booktitle     = {International Conference on Learning Representations (ICLR)},
  year          = {2024},
  url           = {https://arxiv.org/abs/2307.13854}
}

@inproceedings{seer,
  author        = {Ruoyu Qin and Weiran He and Weixiao Huang and Yangkun Zhang and Yikai Zhao and Bo Pang and Xinran Xu and Yingdi Shan and Yongwei Wu and Mingxing Zhang},
  title         = {Seer: Online Context Learning for Fast Synchronous {LLM} Reinforcement Learning},
  booktitle     = {20th USENIX Symposium on Operating Systems Design and Implementation (OSDI 26)},
  year          = {2026},
  url           = {https://arxiv.org/abs/2511.14617}
}

@article{deepseekmath,
  author        = {Zhihong Shao and Peiyi Wang and Qihao Zhu and Runxin Xu and Junxiao Song and Xiao Bi and Haowei Zhang and Mingchuan Zhang and Y. K. Li and Y. Wu and Daya Guo},
  title         = {{DeepSeekMath}: Pushing the Limits of Mathematical Reasoning in Open Language Models},
  journal       = {arXiv preprint arXiv:2402.03300},
  year          = {2024},
  url           = {https://arxiv.org/abs/2402.03300}
}

@article{tvcache,
  author        = {Abhishek Vijaya Kumar and Bhaskar Kataria and Byungsoo Oh and Emaad Manzoor and Rachee Singh},
  title         = {{TVCache}: A Stateful Tool-Value Cache for Post-Training {LLM} Agents},
  journal       = {arXiv preprint arXiv:2602.10986},
  year          = {2026},
  url           = {https://arxiv.org/abs/2602.10986}
}

@article{venusrl,
  author        = {Mingjun Zhang and Yucheng Li and Menghao Zhang and Shuyong Zhu and Ping Zhang and Xiaohe Hu and Jun Chen and Zhixin Wang and others},
  title         = {{VenusRL}: A Fully Disaggregated Agentic {RL} System with Priority Scheduling and Scalable Interaction},
  journal       = {arXiv preprint arXiv:2610.03286},
  year          = {2026},
  url           = {https://arxiv.org/abs/2610.03286}
}

@misc{agentenv,
  author        = {{kvcache-ai}},
  title         = {{AgentENV}: Sandbox Runtime for Agentic {RL}},
  year          = {2026},
  howpublished  = {\url{https://github.com/kvcache-ai/AgentENV}},
  note          = {Accessed: 2026-10-25}
}

@misc{ksm,
  author        = {{The Linux Kernel Documentation}},
  title         = {Kernel Samepage Merging},
  year          = {2026},
  howpublished  = {\url{https://docs.kernel.org/admin-guide/mm/ksm.html}},
  note          = {Accessed: 2026-10-25}
}

@article{treegrpo,
  author        = {Yuxiang Ji and Ziyu Ma and Yong Wang and Guanhua Chen and Xiangxiang Chu and Liaoni Wu},
  title         = {Tree Search for {LLM} Agent Reinforcement Learning},
  journal       = {arXiv preprint arXiv:2509.21240},
  year          = {2025},
  url           = {https://arxiv.org/abs/2509.21240}
}

@article{crab,
  author        = {Tianyuan Wu and Chaokun Chang and Lunxi Cao and Wei Gao and Wei Wang},
  title         = {Crab: A Semantics-Aware Checkpoint/Restore Runtime for Agent Sandboxes},
  journal       = {arXiv preprint arXiv:2604.28138},
  year          = {2026},
  url           = {https://arxiv.org/abs/2604.28138}
}

@article{deltabox,
  author        = {Yunpeng Dong and Jingkai He and Shiqi Liu and Yuze Hou and Dong Du and Zhonghu Xu and Si Yu and Baochuan Yang and others},
  title         = {{DeltaBox}: Scaling Stateful {AI} Agents with Millisecond-Level Sandbox Checkpoint/Rollback},
  journal       = {arXiv preprint arXiv:2605.22781},
  year          = {2026},
  url           = {https://arxiv.org/abs/2605.22781}
}

@article{cuasandbox,
  author        = {Xin Yan and Zhengbo Jiao and Jiaqi Liu and Zhenglin Wan and SiYuan Ma and Xuliang Yu and Tianyi Jiang and Chubin Zhang and others},
  title         = {{CUA-Sandbox}: Efficient Environments for Computer-Use Agent Reinforcement Learning},
  journal       = {arXiv preprint arXiv:2609.32750},
  year          = {2026},
  url           = {https://arxiv.org/abs/2609.32750}
}

@article{rmcts,
  author        = {Xiao Yu and Baolin Peng and Vineeth Vajipey and Hao Cheng and Michel Galley and Jianfeng Gao and Zhou Yu},
  title         = {{ExACT}: Teaching {AI} Agents to Explore with Reflective-{MCTS} and Exploratory Learning},
  journal       = {arXiv preprint arXiv:2410.02052},
  year          = {2024},
  url           = {https://arxiv.org/abs/2410.02052}
}

@article{qwen3vl,
    author        = {{Qwen Team}},
    title         = {{Qwen3-VL} Technical Report},
    journal       = {arXiv preprint arXiv:2511.21631},
    year          = {2025},
    url           = {https://arxiv.org/abs/2511.21631}
  }

@inproceedings{catalyzer,
  author        = {Dong Du and Tianyi Yu and Yubin Xia and Binyu Zang and Guanglu Yan and Chenggang Qin and Qixuan Wu and Haibo Chen},
  title         = {Catalyzer: Sub-millisecond Startup for Serverless Computing with Initialization-less Booting},
  booktitle     = {Proceedings of the 25th International Conference on Architectural Support for Programming Languages and Operating Systems (ASPLOS)},
  year          = {2020},
  doi           = {10.1145/3373376.3378512}
}

@inproceedings{medes,
  author        = {Divyanshu Saxena and Tao Ji and Arjun Singhvi and Junaid Khalid and Aditya Akella},
  title         = {Memory Deduplication for Serverless Computing with {Medes}},
  booktitle     = {Proceedings of the Seventeenth European Conference on Computer Systems (EuroSys)},
  year          = {2022},
  pages         = {714--729},
  doi           = {10.1145/3492321.3524272}
}

@inproceedings{mitosis,
  author        = {Xingda Wei and Fangming Lu and Tianxia Wang and Jinyu Gu and Yuhan Yang and Rong Chen and Haibo Chen},
  title         = {No Provisioned Concurrency: Fast {RDMA}-codesigned Remote Fork for Serverless Computing},
  booktitle     = {17th USENIX Symposium on Operating Systems Design and Implementation (OSDI 23)},
  year          = {2023},
  pages         = {497--517}
}

@article{statefork,
  author        = {Jiakai Xu and Tianle Zhou and Georgios Liargkovas and Danielle Gillai and Ruizhe Fu and Patrick Shen and Eugene Wu and Kostis Kaffes},
  title         = {{StateFork}: Branchable Infrastructure for Agent Exploration},
  journal       = {arXiv preprint arXiv:2609.38648},
  year          = {2026},
  url           = {https://arxiv.org/abs/2609.38648}
}

@article{paste,
  author        = {Yifan Sui and Han Zhao and Rui Ma and Zhiyuan He and Hao Wang and Jianxun Li and Kaiqiang Xu and Kai Chen and others},
  title         = {Parallelizing Tool Execution and {LLM} Generation for Low-Latency Agent Serving},
  journal       = {arXiv preprint arXiv:2603.18897},
  year          = {2026},
  url           = {https://arxiv.org/abs/2603.18897}
}

@article{agenticrlsurvey,
  author        = {Guibin Zhang and Hejia Geng and Xiaohang Yu and Zhenfei Yin and Zaibin Zhang and Zelin Tan and Heng Zhou and Zhongzhi Li and others},
  title         = {The Landscape of Agentic Reinforcement Learning for {LLMs}: A Survey},
  journal       = {Transactions on Machine Learning Research},
  year          = {2026},
  url           = {https://arxiv.org/abs/2509.02547}
}

@article{toolcaching,
  author = {Yi Zhai and Dian Shen and Junzhou Luo and Bin Yang},
  title = {{ToolCaching}: Towards Efficient Caching for {LLM} Tool-calling},
  journal = {arXiv preprint arXiv:2601.15335},
  year = {2026},
  url = {https://arxiv.org/abs/2601.15335}
}

@article{nichols2025speculative,
  author = {Daniel Nichols and Prajwal Singhania and Charles Jekel and Abhinav Bhatele and Harshitha Menon},
  title = {Optimizing Agentic Language Model Inference via Speculative Tool Calls},
  journal = {arXiv preprint arXiv:2512.15834},
  year = {2025},
  url = {https://arxiv.org/abs/2512.15834}
}

@article{huang2025searchspec,
  author = {Zixiao Huang and Wen Zeng and Tianyu Fu and Tengxuan Liu and Yizhou Sun and Ke Hong and Xinhao Yang and Chengchun Liu and Yan Li and Quanlu Zhang and Guohao Dai and Zhenhua Zhu and Yu Wang},
  title = {Reducing Latency of {LLM} Search Agent via Speculation-based Algorithm-System Co-Design},
  journal = {arXiv preprint arXiv:2511.20048},
  year = {2025},
  url = {https://arxiv.org/abs/2511.20048}
}

@article{speculativeactions,
  author = {Naimeng Ye and Arnav Ahuja and Georgios Liargkovas and Yunan Lu and Kostis Kaffes and Tianyi Peng},
  title = {Speculative Actions: A Lossless Framework for Faster Agentic Systems},
  journal = {arXiv preprint arXiv:2510.04371},
  year = {2025},
  url = {https://arxiv.org/abs/2510.04371}
}

@article{gupta2026cache,
  author = {Shivam Gupta},
  title = {Marginally Correct Tool Caches Can Reverse Group-Normalized Policy Updates},
  journal = {arXiv preprint arXiv:2609.26866},
  year = {2026},
  url = {https://arxiv.org/abs/2609.26866}
}
\bibliographystyle{mlsys2025}

\end{document}